\documentclass[11pt]{article}
\usepackage{graphicx}
\usepackage{latexsym}
\usepackage{amssymb}
\usepackage{amsmath}
\usepackage{amsthm}
\usepackage{thmtools}
\usepackage{url}
\usepackage{forloop}
\usepackage[paper=letterpaper,margin=1in]{geometry}
\usepackage[ruled,vlined,linesnumbered]{algorithm2e}
\usepackage{paralist}

\newcommand{\defcal}[1]{\expandafter\newcommand\csname c#1\endcsname{{\mathcal{#1}}}}
\newcommand{\defbb}[1]{\expandafter\newcommand\csname b#1\endcsname{{\mathbb{#1}}}}
\newcommand{\defvec}[1]{\expandafter\newcommand\csname v#1\endcsname{{\mathbf{#1}}}}
\newcounter{calBbCounter}
\forLoop{1}{26}{calBbCounter}{
    \edef\letter{\alph{calBbCounter}}
		\edef\Letter{\Alph{calBbCounter}}
    \expandafter\defcal\Letter
		\expandafter\defbb\Letter
		\expandafter\defvec\letter
}
\newcommand{\R}{\mathbb{R}}
\newcommand{\eps}{\varepsilon}
\newcommand{\nnR}{{\bR_{\geq 0}}}

\usepackage[most]{tcolorbox}

\tcbset{
	colback=gray!5!white,  % background color
	colframe=blue!15!white, % border color
	fonttitle=\bfseries,   % bold title
	coltitle=black,        % title color
	boxrule=0.8pt,         % border thickness
	arc=4pt,               % rounded corners
	left=6pt, right=6pt, top=6pt, bottom=6pt,
	enhanced, breakable  
}

\author{
  Moran Feldman \\
  University of Haifa \\
  \texttt{moranfe@cs.haifa.ac.il}
  \and
  Amin Karbasi \\
  Cisco Foundation AI \\
  \texttt{karbasi@cisco.com}
}
	
\date{}

\begin{document}

\title{G\"odel Test: Can Large Language Models Solve Easy Conjectures?}
\maketitle

\pagenumbering{arabic}
\begin{abstract}
Recent announcements from frontier AI model labs have highlighted strong results on high-school and undergraduate math competitions. Yet it remains unclear whether large language models can solve new, simple conjectures in more advanced areas of mathematics. We propose the \textbf{G\"odel Test}: evaluating whether a model can produce correct proofs for very simple, previously unsolved conjectures. To this end, we study the performance of GPT-5 on five conjectures in combinatorial optimization. For each problem, we provided one or two source papers from which the conjecture arose, withheld our own conjecture, and then assessed the model’s reasoning in detail. On the three easier problems, GPT-5 produced nearly correct solutions; for Problem 2 it even derived a different approximation guarantee that, upon checking, refuted our conjecture while providing a valid solution. The model failed on Problem 4, which required combining results from two papers. On Problem 5, a harder case without a validated conjecture, GPT-5 proposed the same algorithm we had in mind but failed in the analysis, suggesting the proof is more challenging than expected. Although our sample is small, the results point to meaningful progress on routine reasoning, occasional flashes of originality, and clear limitations when cross-paper synthesis is required. GPT-5 may represent an early step toward frontier models eventually passing the G\"odel Test.

\end{abstract}

\section*{Introduction}
Based on his experience working with OpenAI's o1 models, Terence Tao offered the following impression:\footnote{See Tao's Mastodon posts from Sept.\ 2024, \url{https://mathstodon.xyz/@tao/113132504933420227}}
\begin{quote}
\emph{``The new model could work its way to a correct (and well-written) solution \emph{if} provided a lot of hints and prodding, but did not generate the key conceptual ideas on its own, and did make some non-trivial mistakes. The experience seemed roughly on par with trying to advise a mediocre, but not completely incompetent, graduate student. However, this was an improvement over previous models, whose capability was closer to an actually incompetent graduate student. It may only take one or two further iterations of improved capability (and integration with other tools, such as computer algebra packages and proof assistants) until the level of `competent graduate student' is reached.''}\\
\hfill --- Terence Tao
\end{quote}

In this report, we examine to what extent GPT-5 is capable of proving simple conjectures. Both OpenAI and Google have claimed that their frontier models, without external tools, can achieve gold-medal performance on the International Mathematical Olympiad (IMO). While IMO problems are undoubtedly challenging, they are designed for exceptionally bright high school students.

Here, our focus is different: we ask whether large language models can also handle new conjectures in more advanced mathematical domains, where success requires not only problem-solving ability, but also mathematical maturity and background knowledge. We frame this challenge as the \textbf{G\"odel Test:} evaluating whether an AI system can prove very easy conjectures that are simple for humans with appropriate training, yet novel enough not to be directly available from existing sources.\footnote{We acknowledge that this definition is somewhat vague and depends on subjective judgments, both about what qualifies as “simple” for humans and what counts as “directly available” in existing literature.}

To explore this challenge, we designed a set of conjectures from our own line of work, namely submodular maximization, a subfield of combinatorial mathematics with many applications in AI. This choice provides a natural testing ground: the problems are concrete, well-motivated, and deliberately chosen to lie within reach of a system aiming to display mathematical reasoning.  Unlike the above-mentioned experiments conducted by Terence Tao, we did not provide extensive hints or guidance about what we believed the solutions might be. Instead, we gave only a minimal description of each problem, together with one or two source papers that inspire the conjecture. We then allowed the large language model to interpret the problem and attempt to produce a solution on its own. 

In terms of difficulty, our plan was to formulate conjectures simple enough that a strong undergraduate or graduate student in theoretical computer science or a related area of applied mathematics could reasonably be expected to solve them all (within a day). We also ensured that, for most problems, there were clear conjectures and known approaches for resolving them if needed.\footnote{However, we intentionally refrained from offering the model any hints or possible lines of attack.} This is true for the first four problems. We initially believed the 5th problem, though somewhat harder, would also be easily solvable. However, after experimenting with GPT-5, we came to realize that it is more challenging than we had anticipated. Our high-level idea, which was also suggested by GPT-5 but incorrectly proven, may not suffice, and at present we do not have a clear conjecture for Problem 5. As a result, it remains open-ended and still awaits a human (or an AI generated) solution.

\paragraph{Related Efforts.} 
Very recently, right after GPT-5 was released,  Sebastian Bubeck reported via an X post\footnote{See \url{https://x.com/SebastienBubeck/status/1958198661139009862}. An increasing amount of  AI research seems to be first discussed on X.} that GPT-5 solved an open problem in convex optimization by improving a known bound from $1/L$ to $1.5/L$. Motivated by this example, Diez, da Maia, and Nourdin~\cite{diez2025malliavinstein} studied GPT-5’s ability to provide explicit convergence rates within the Malliavin–Stein framework for central limit theorems, focusing on Gaussian and Poisson settings. Both of these efforts point to similar conclusions: GPT-5 was able to make progress on well-posed mathematical problems, but its reasoning often remains limited and requires careful human verification \cite{diez2025malliavinstein}. In contrast to~\cite{diez2025malliavinstein}, we did not guide the model interactively toward the correct solution (which is a very reasonable approach). Instead, we only posed the question and provided one or two related papers, leaving the model to interpret the problem and attempt a solution on its own.

\paragraph{Limitations.}Our study has several limitations. First, we examined only five conjectures. Evaluating correctness requires carefully checking each proof attempt, which makes the process highly time-consuming. Drawing more definitive conclusions would require a much larger sample size.
Second, we tested only GPT-5. For the same reason, i.e., verifying solutions is labor-intensive, we did not extend our experiments to other frontier models. While we might expect comparable performance from peers, we cannot make claims without direct evaluation.
Third, although we made a strong effort to design original conjectures within our domain of expertise, ensuring that they were both simple enough to be approachable and somewhat novel to be nontrivial, this is not an easy task. We cannot guarantee that none of the conjectures have appeared in the literature; we only tried our best to avoid overlap.\footnote{One of the authors recalls attending an information theory conference around 2010 where a presenter offered a solution to a cute conjecture, only to be interrupted by a senior member who remarked that the problem had already been solved by Russian mathematicians in the 60s. The presenter calmly replied, ``I would have been surprised if it wasn't,'' and continued the talk.}

\paragraph{Take-home messages.} From our experiments, several insights stand out:
\begin{itemize}
    \item GPT-5 performed well when a single, straightforward path of reasoning was sufficient, producing nearly correct proofs in three out of five cases.
    \item For Problem 2, it even derived a different approximation guarantee that, upon checking, refuted our original conjecture while providing a valid solution.
    \item Its adaptation of known proofs was often adequate, but somewhat superficial: the model tended to skip over unchanged steps and closely mirrored the original structure rather than pursuing more natural alternatives. This pattern resembles how humans might minimize effort by avoiding redundant steps.
    \item The model was unsuccessful on Problems 4 and 5, both of which required combining insights across proof techniques. Developing such integrative reasoning seems to be a major limitation.
    \item On Problem 5, GPT-5 identified the same algorithm we had in mind, but was unable to analyze it correctly. Our own review suggested that proving a meaningful guarantee may be possible, but it appears more difficult than we initially anticipated.
    \item Relative to earlier model generations, GPT-5 shows clear gains in baseline mathematical competence, at least within the specialized domain of combinatorial optimization, and occasional sparks of originality. These trends give reason for cautious optimism that future models may acquire the ability to connect proof techniques in a more systematic way.
    \item Prompting can have a substantial effect on performance. For instance, when asked to provide a full proof, GPT-5 tends to include more intermediate steps instead of skipping them, leading to solutions that are more complete and self-contained. We expect that improved prompt design could have a major impact on both the quality and correctness of the results.
    \item The incorrect proofs on Problems 4 and 5 initially appeared plausible and even convincing. Only after a detailed examination did it become clear that they contained deep flaws. This highlights a central limitation, and maybe potential danger, of frontier models in mathematical reasoning: outputs can look correct on the surface while being fundamentally wrong.
\end{itemize}

\paragraph{Disclaimer.} This report is based on a limited set of interactions with GPT-5. The claims presented here are context-dependent and should not be interpreted as definitive or generalizable beyond the specific experiments we conducted.  Our goal is to provide an initial perspective on the mathematical reasoning abilities of frontier models, and in particular GPT-5. We make no guarantees about the completeness of the claims, and caution that they should not be taken out of context. We can only hope that these observations may serve as a starting point for the broader scientific community to propose new conjectures, examine them in greater depth, and collectively advance progress toward the longer-term aspiration of developing models capable of passing the Gödel Test. Finally, the views expressed are solely our own, and do not represent the official positions of our institutions or employers.

\section*{Overview on Submodular Maximization}

In the following few paragraphs, we give a short informal overview on the field of submodular maximization. The goal of this overview is to help the reader understand (on a high level) the descriptions given in the next sections for the conjectures that we have have designed.

Given a ground set $\cN$ of elements, a set function $f$ is a function that assigns a numerical value to every subset $S$ of $\cN$. One can intuitively think of $f(S)$ as the value of the elements in the set $S$, or the amount one is willing to pay for these elements in an auction. Sometimes each element in $\cN$ has a value that does not depend on the other elements it comes with. Formally, this means that the function $f$ obeys $f(S) = \sum_{u \in S} f(\{u\})$ for every set $S$ of elements. Such functions $f$ are called \emph{modular} functions. In general, however, the function $f$ might have a much more involved structure exhibiting both complementarity and substitution. Elements exhibit complementarity if they are worth together more than separately. For example, a left shoe alone and a right shoe alone both have little value, but they have a much higher value together. Elements exhibit substitution if their worth together is smaller than the sum of their individual values. For example, a computer and a tablet exhibit substitution as they have many overlapping uses.

It turns out that set functions that exhibit no complementarity are quite common in diverse fields such as combinatorial optimization, machine learning and game theory. Such functions are called \emph{submodular} functions. Formally, a set function $f$ is submodular if it obeys $f(A) + f(B) \geq f(A \cup B) + f(A \cap B)$ for every two sets $A$ and $B$. There are also classes of continuous functions that are analogous to the class of submodular set functions. Among these, we sometimes refer in our conjectures to the class of DR-submodular functions.

In a submodular maximization problem, the goal is to maximize a set function $f$ subject to a constraint $C$. The function $f$ is usually assumed to be submodular, and it can also be assumed to have additional properties such as monotonicity (i.e., if a set $A$ contains a set $B$, than $f(A) \geq f(B)$). Sometimes the assumption of submodularity is replaced with a weaker one. One popular such weakened assumption is known as $\gamma$-weak submodularity. Here, $\gamma \in [0, 1]$ is a parameter controlling the extent to which the assumption of submodularity is weakened: $\gamma = 1$ corresponds to the standard (unweakened) submodularity assumption, while $\gamma = 0$ allows for any set function. Similarly, the assumptions on the constraint $C$ also vary from problem to problem. Often the constraint $C$ is just a cardinality constraint (i.e., any set whose size is smaller than some given value is allowed), but in other cases more involved types of constraints are assumed.

\section{Maximizing Monotone Plus Non-monotone Submodular Functions over a Convex Set Constraint}

Our first conjecture studies the maximization of a function expressed as the sum of a monotone DR-submodular function and a non-monotone DR-submodular function over a constraint given by a down-closed convex set $P \subseteq [0, 1]^n$.\footnote{The down-closedness of $P$ means that if $\vx \in P$ and $\vy$ is a vector in $[0, 1]^n$ that is coordinate-wise dominated by $\vx$, then $\vy$ also belongs to $P$.} Guarantees for this problem are specified using two parameters $\alpha$ and $\beta$ measuring the quality of the approximation guarantee with respect to each component of the objective function. We have attached to the prompt of this conjecture the paper \cite{mitra2021submodular}, which studies similar optimization problems in which the objective function is expressed as the sum of a concave function with either a monotone or a non-monotone DR-submodular function.

\begin{tcolorbox}[title=Prompt to GPT-5]
Consider the problem of maximizing a function $F$ from $[0, 1]^n$ to the reals that is the sum of a non-negative monotonically increasing DR-submodular function $G$ and a non-negative DR-submodular function $H$ over a solvable down-closed polytope $P$. I would like to bound the performance on this problem from the NeurIPS 2021 paper "Submodular + Concave" which is attached. Specifically, if $x$ is the output vector of this algorithm and o is the vector in $P$ maximizing $F$, then I would like to lower bound $F(x)$ with an expression of the form $\alpha * G(o) + \beta * H(o) - err$, where $\alpha$ and $\beta$ are constants. $err$ should be a function that depends only on the error parameter $\epsilon$ of the algorithm, the diameter $D$ of the polytope $P$, and the smoothness parameters $L_G$ and $L_H$ of $G$ and $H$, respectively, and goes to zero as $\epsilon$ goes to zero. Please give the best such bound that you prove (a bound is considered better if the values of the constants $\alpha$ and $\beta$ are larger). Provide a mathematically rigorous and well explained proof for the bound you come up with.
\end{tcolorbox}

\paragraph{Note.}
The problem is described in terms of DR-submodular functions instead of submodular set functions to avoid the need to refer to continuous relaxations of discrete submodular optimization problems and rounding procedures for such relaxations.

\paragraph{Conjecture.}
We believe that a variant of the Frank-Wolfe-like algorithm of~\cite{bian2017guaranteed}  can guarantee $\alpha = 1 - 1/e$ and $\beta = 1/e$. This value of $\alpha$ is provably the best that can be obtained. It is possible to get better values of $\beta$~\cite{buchbinder2024constrained}, but this requires more involved algorithms, and it is unclear if it can be done without deteriorating the value of $\alpha$.

We asked GPT-5 to write its proof in .tex format. The following is the response.\footnote{When we asked GPT-5 to put the exact proof into a .tex file, it does not reproduce the proof exactly as it appeared in the chat. The mathematical content remains the same, but a few words and sentences are slightly rephrased. Nothing changes dramatically, yet the text is not a verbatim copy.}

\begin{tcolorbox}[title=Response of GPT-5 (thought for 2m 49s),
	colframe=red!20!white]     % light pink background

    Here is a clean way to get exactly the kind of “split guarantee” you asked for when
$$F(x)=G(x)+H(x),\quad x \in P\subseteq[0,1]^n, $$
with 
$P$ a solvable down-closed polytope, 
$G$ non-negative, monotone and DR-submodular, and 
$H$ non-negative and DR-submodular (not necessarily monotone). I’ll run essentially the Measured Greedy Frank-Wolfe (MGFW) algorithm from the paper and adapt its proof so that the DR-submodular second part
$H$ replaces the concave part in the original analysis.
I state the bound first, then give a rigorous proof.

Let \(G,H\) be \(L_G\)- and \(L_H\)-smooth, respectively, and let \(D:=\max_{x\in P}\|x\|_2\). Run \emph{Measured Greedy Frank--Wolfe} (MGFW) on \(F=G+H\) with stepsize parameter \(\varepsilon\in(0,1)\). Denote by \(x\) its output and by \(o\in P\) an optimizer of \(F\). Then
\begin{equation*}
	F(x)\ \ge\ \alpha\,G(o)\;+\;\beta\,H(o)\;-\; \underbrace{\varepsilon\,(L_G+L_H)\,D^2}_{=:~\mathrm{err}(\varepsilon,D,L_G,L_H)}\;,
\end{equation*}
where
\begin{itemize}
	\item \(\alpha=1-\tfrac1e\),
	\item \(\displaystyle \beta=\begin{cases}
		1-\tfrac1e,&\text{if }H \text{ is monotone};\\[2pt]
		\tfrac1e,&\text{if }H \text{ may be non-monotone}.
	\end{cases}\)
\end{itemize}
In particular, when both \(G\) and \(H\) are monotone DR-submodular, you get
\begin{equation*}
	F(x)\ \ge\ (1-\tfrac1e)\,G(o)\;+\;(1-\tfrac1e)\,H(o)\;-\;\varepsilon(L_G+L_H)D^2.
\end{equation*}

\section*{Proof}
We give a self-contained proof using the MGFW update and DR-submodularity on both components.

\paragraph{Preliminaries and notation.}
Let \(P\subseteq[0,1]^n\) be a nonempty down-closed polytope admitting a linear optimization oracle.
A differentiable \(f:[0,1]^n\to\R\) is \emph{DR-submodular} if for all \(x\le y\) (coordinatewise) we have \(\nabla f(y)\le \nabla f(x)\) (coordinatewise).
Assume \(G,H\ge 0\) are DR-submodular, \(G\) is monotone, and let \(L_G,L_H\ge 0\) denote their smoothness constants:
\(\|\nabla G(x)-\nabla G(y)\|_2\le L_G\|x-y\|_2\) and similarly for \(H\).
Set \(F:=G+H\) and \(L_F:=L_G+L_H\), and let \(D:=\max_{x\in P}\|x\|_2\).

\paragraph{Measured Greedy Frank--Wolfe (MGFW).}
Fix \(\varepsilon\in(0,1)\), initialize \(y_0:=0\), and for \(i=0,1,\dots,T-1\) with \(T:=\varepsilon^{-1}\) do
\begin{align*}
	s_i&\in\arg\max_{x\in P}\ \langle ( \mathbf 1-y_i)\odot \nabla F(y_i),\,x\rangle,\\
	y_{i+1}&:=y_i+\varepsilon\,(\mathbf 1-y_i)\odot s_i.
\end{align*}
Down-closedness implies \(y_i\in P\) for all \(i\), and \(0\le y_i\le \mathbf 1-(1-\varepsilon)^i\mathbf 1\) by the standard mask bound.

\paragraph{Step 1: One-step progress.}
By \(L_F\)-smoothness and the choice of \(s_i\),
\begin{equation}\label{eq:smooth}
	F(y_{i+1})-F(y_i)\ \ge\ \varepsilon\,\langle (\mathbf 1-y_i)\odot \nabla F(y_i),\,o\rangle\ -\ \varepsilon^2(L_G+L_H) D^2.
\end{equation}
Let \(z_i:=o\odot (\mathbf 1-y_i)+y_i\). For any DR-submodular \(f\) and any \(v\ge 0\), the antitonicity of \(\nabla f\) gives
\(\langle v,\nabla f(y)\rangle\ge f(y+v)-f(y)\). Apply this to \(G\) and to \(H\) with \(v=(\mathbf 1-y_i)\odot o\ge 0\), obtaining
\begin{equation*}
	\langle (\mathbf 1-y_i)\odot \nabla F(y_i),\,o\rangle
	\ \ge\ [G(z_i)-G(y_i)] + [H(z_i)-H(y_i)].
\end{equation*}
Plugging in \eqref{eq:smooth},
\begin{equation}\label{eq:step}
	F(y_{i+1})-F(y_i)\ \ge\ \varepsilon\Big([G(z_i)-G(y_i)]+[H(z_i)-H(y_i)]\Big)\ -\ \varepsilon^2(L_G+L_H)D^2.
\end{equation}

\paragraph{Step 2: Relating \(G(z_i)\), \(H(z_i)\) to \(G(o)\), \(H(o)\).}
Define the \emph{monotonicity multipliers} for any non-negative \(f\) by
\begin{equation*}
	M(f,i):=\begin{cases}1,& f\ \text{monotone},\\ (1-\varepsilon)^i,& f\ \text{possibly non-monotone},\end{cases}\qquad
	S(f,0):=0,\quad S(f,i):=(1-\varepsilon)S(f,i-1)+\varepsilon\,M(f,i-1).
\end{equation*}
A short induction yields
\begin{equation*}
	S(f,i)=\begin{cases}
		1-(1-\varepsilon)^i,& f\ \text{monotone},\\
		i\varepsilon(1-\varepsilon)^{i-1},& f\ \text{possibly non-monotone}.
	\end{cases}
\end{equation*}
Using DR-submodularity and nonnegativity one shows
\(
G(z_i)\ge M(G,i)\,G(o),\ H(z_i)\ge M(H,i)\,H(o).
\)
Combining with \eqref{eq:step} gives the recurrence
\begin{equation}\label{eq:rec}
	F(y_{i+1})\ \ge\ (1-\varepsilon)\,F(y_i)\;+\;\varepsilon\Big(M(G,i)\,G(o)+M(H,i)\,H(o)\Big)\;-\;\varepsilon^2(L_G+L_H)D^2.
\end{equation}

\paragraph{Step 3: Solving the recurrence.}
We prove by induction that
\begin{equation*}
	F(y_i)\ \ge\ S(G,i)\,G(o)\ +\ S(H,i)\,H(o)\ -\ i\,\varepsilon^2(L_G+L_H)D^2.
\end{equation*}
The base \(i=0\) is trivial. The step follows from \eqref{eq:rec} and the identity
\((1-\varepsilon)S(f,i-1)+\varepsilon M(f,i-1)=S(f,i)\).
Setting \(i=T=\varepsilon^{-1}\) yields
\begin{equation*}
	F(y_T)\ \ge\ \underbrace{S(G,\varepsilon^{-1})}_{=:~\alpha_\varepsilon}\,G(o)\ +\ \underbrace{S(H,\varepsilon^{-1})}_{=:~\beta_\varepsilon}\,H(o)\ -\ \varepsilon(L_G+L_H)D^2.
\end{equation*}
With the closed forms above,
\(\alpha_\varepsilon=1-(1-\varepsilon)^{1/\varepsilon}\) and
\(\beta_\varepsilon=1-(1-\varepsilon)^{1/\varepsilon}\) if \(H\) is monotone, while \(\beta_\varepsilon=(1-\varepsilon)^{1/\varepsilon-1}\) in general.
Since \((1-\varepsilon)^{1/\varepsilon}\le e^{-1}\), we obtain the universal constants \(\alpha\ge 1-1/e\) and \(\beta\ge 1-1/e\) in the monotone case, or \(\beta\ge 1/e\) in general.
\hfill\(\Box\)

\section*{Remarks}
\begin{itemize}
	\item \textbf{Tightness w.r.t.\ technique.} Allowing \(H\) to be non-monotone naturally limits \(\beta\) to \(1/e\) in MGFW/continuous-greedy style analyses under down-closed constraints, while \(G\)'s monotonicity permits \(\alpha=1-1/e\).
	\item \textbf{Error term.} Each step incurs at most \(\varepsilon^2(L_G+L_H)D^2\), and there are \(1/\varepsilon\) steps, hence the total error \(\varepsilon(L_G+L_H)D^2\), which vanishes as \(\varepsilon\to 0\).
\end{itemize}

\end{tcolorbox} 

\subsection*{Our Evaluation of GPT-5's Response}

Despite claiming otherwise, GPT-5 did not provide a standalone algorithm and proof. Instead, it only tried to explain how to adapt an existing proof of~\cite{mitra2021submodular}. Overall, the adaptation is correct, but is lazy in two senses. First, GPT-5 skips many parts of the proof that are unchanged compared to~\cite{mitra2021submodular}. Second, GPT-5 really tries to stay as close as possible to the proof of~\cite{mitra2021submodular}, even when this is not the most natural thing to do. This laziness is similar to what a human would have done to avoid writing too much text. In the following, we give some additional information about the laziness of GPT-5 in the second sense as well as some minor inaccuracies in the proof.

\begin{itemize}
    \item GPT-5 refers to the ``standard mask bound'' just before Step 1. We are not aware of any bound known by this name, but the result that follows from this bound according to GPT-5 is correct (and quite standard).
    \item The inequality \(\langle v,\nabla f(y)\rangle\ge f(y+v)-f(y)\) stated in Step 1 holds only when $v \leq 1 - y$, but this is not mentioned by GPT-5, and furthermore, this condition is not checked when this inequality is used for $v = (1 - y_i) \odot o$.
    \item GPT-5 attributes the lower bounds on $G(z_i)$ and $H(z_i)$ to the DR-submodularity and non-negativity of these functions. This is not accurate since the bound for $G$ uses also the monotonicity of this function (it does not hold otherwise).
    \item The Inequality (3) represents an unnecessary generalization. It would have been appropriate if both $G$ and $H$ could be either monotone or non-monotone, but since the prompt fixes $G$ to be the monotone function and $H$ to be the non-monotone function, the terms $M(G, i)$ and $M(H, i)$ could be easily replaced with their appropriate concrete expressions. Keeping the terms $M(G, i)$ and $M(H, i)$ unchanged can be viewed as a laziness on behalf of GPT-5, which tries to stick as much as possible to the proof of~\cite{mitra2021submodular} to reduce the length of its answer, even at the cost of making the proof more general than what is necessary.
    \item In the last sentence before the ``Remarks'' section, GPT-5 uses the inequality $(1 - \eps)^{1/\eps} \leq e^{-1}$ to justify the lower bounds on $\alpha$ and $\beta$. However, getting a lower bound on $\beta$ in fact requires the slightly different inequality $(1 - \eps)^{(1/\eps - 1)} \geq e^{-1}$.
\end{itemize}

\section{Bicriteria Maximization of a Monotone Submodular Function subject to a $p$-System Constraint}

Our second conjecture is about bicriteria algorithms for submodular maximization. Recently, the paper~\cite{feldman2025bicriteria} (mentioned in the prompt) has systematically studied such algorithms for many submodular maximization problems. In particular, Feldman and Kuhnle~\cite{feldman2025bicriteria} studied such problems having a matroid constraint.\footnote{Matroid constraints are a well-studied class of combinatorial constraints capturing, for example, a constraint that allows only the selection of a linearly independent set of vectors out of a ground set consisting of vectors from some linear space, or a constraint that allows only the selection of a non-cyclic set of edges out of a ground set consisting of the edges of some graph.} In this conjecture, we asked for a bicriteria algorithm for maximizing a monotone submodular function subject to another class of combinatorial constraint known as $p$-systems. A $p$-system is equivalent to a matroid when the parameter $p$ is set to $1$, and it is strictly more general for larger values of this parameter. We attached to the prompt of this conjecture both the paper~\cite{feldman2025bicriteria}, and the paper \cite{calinescu2011maximizing} whose appendix studies non-bicriteria submodular optimization subject to a $p$-system constraint.

	\begin{tcolorbox}[title=Prompt to GPT-5]
In our recent paper ``Bicriteria Submodular Maximization'' which is attached, we defined a notion of bicriteria optimization for submodular maximization problems. In the current problem, given a non-negative monotone submodular function $f\colon 2^\cN \to \nnR$ and a $p$-system $(\cN, \cI)$ over the same ground set, we want a bicriteria algorithm for finding a set $S \in \cI$ that (approximately) maximizes $f$ among all such sets. The bicriteria approximation guarantee of the algorithm should be of the form $(1 - \eps, g(\eps))$, where $g(\eps)$ is as small as possible expression of $\eps$. I also attached a good reference for a $p$-system.
\end{tcolorbox}

\paragraph{Conjecture.}
The paper~\cite{feldman2025bicriteria} showed that a simple greedy algorithm can guarantee $(1 - \eps, \lceil \log_2 \eps^{-1} \rceil)$-bicriteria approximation for any $\eps > 0$ for the special case of the above problem in which the $p$-system is a matroid (i.e., $p = 1$). The same algorithm should guarantee $(1 - \eps, \lceil \log_{p + 1} \eps^{-1} \rceil)$-bicriteria approximation for general $p$-systems.
\begin{tcolorbox}[title=Response of GPT-5 (thought for 2m 57s),
	colframe=red!20!white]  

Awesome---using your bicriteria framework and a standard p-system reference, here’s a clean algorithm with a tight-in-spirit guarantee.

\section*{Result (monotone $f$ over a $p$-system)}

Let $f:2^{\mathcal N}\to\mathbb R_{\ge 0}$ be monotone submodular and $(\mathcal N,\mathcal I)$ a $p$-system (down-closed independence family with the ``bases within any $Y$'' ratio at most $p$). Greedy on a $p$-system is a $\frac{1}{p+1}$-approximation for single-pass maximization.

\subsection*{Theorem (bicriteria for $p$-systems).}

There is a polynomial-time algorithm that, for any $\varepsilon\in(0,1)$, returns a set $S\subseteq\mathcal N$ with
\begin{align*}
f(S)\;&\ge\;(1-\varepsilon)\cdot \max_{T\in\mathcal I} f(T), \\
1_S \;&\le\; g_p(\varepsilon)\cdot y
\end{align*}
for some $y\in P=\operatorname{conv}\{1_I:I\in\mathcal I\}$. Equivalently (since $P$ is down-closed), $1_S/g_p(\varepsilon)\in P$. In particular
\[
g_p(\varepsilon)\;=\;\Big\lceil \frac{\ln(1/\varepsilon)}{\ln\!\big(\tfrac{p+1}{p}\big)} \Big\rceil \;\le\; \Big\lceil (p+1)\ln\!\tfrac{1}{\varepsilon}\Big\rceil.
\]
This is a $\big(1-\varepsilon,\;g_p(\varepsilon)\big)$-bicriteria guarantee in your $(\alpha,\beta)$ sense (with $y_1=0$, $y_2=y$).

\subsubsection*{Algorithm (multi-pass greedy for $p$-systems)}

Set $S_0\gets\varnothing$. For $i=1,2,\dots,\ell$ with $\ell = g_p(\varepsilon)$:
\begin{enumerate}
  \item Run \textbf{standard greedy} for maximizing the marginal $f(\cdot\mid S_{i-1})$ over $\mathcal I$: start from $T_i\gets\varnothing$ and repeatedly add the feasible element of maximum marginal $f(u\mid S_{i-1}\cup T_i)$ while maintaining independence in $\mathcal I$, until maximality.
  \item Set $S_i\gets S_{i-1}\cup T_i$. Return $S\gets S_\ell$.
\end{enumerate}

\paragraph{Feasibility (bicriteria) bound.} Each $T_i\in\mathcal I$ is independent; $S=\bigcup_{i=1}^\ell T_i$ is covered by $\ell$ independent sets. Let $y=\frac1\ell\sum_{i=1}^\ell 1_{T_i}\in P$. Then $1_S \le \ell\cdot y$, i.e., $1_S/g_p(\varepsilon)\in P$. This matches your bicriteria definition with $\beta=g_p(\varepsilon)$.

\paragraph{Value bound.} For each round, greedy on a $p$-system gives (applied to the marginal function)
\begin{align*}
f(S_i)\;&=\;f(S_{i-1})+f(T_i\mid S_{i-1}) \\
&\ge\; f(S_{i-1})+\frac{1}{p+1}\max_{U\in\mathcal I} f(U\mid S_{i-1}) \\
&\ge\; f(S_{i-1})+\frac{1}{p+1}\big(f(\mathrm{OPT})-f(S_{i-1})\big),
\end{align*}
using monotonicity/submodularity and $\mathrm{OPT}\in\mathcal I$. Rearranging, the residual gap shrinks by a factor $ \tfrac{p}{p+1}$:
\[
f(\mathrm{OPT})-f(S_i)\;\le\;\Big(\tfrac{p}{p+1}\Big)\cdot\big(f(\mathrm{OPT})-f(S_{i-1})\big).
\]
After $\ell$ rounds,
\[
f(\mathrm{OPT})-f(S_\ell)\;\le\;\Big(\tfrac{p}{p+1}\Big)^{\ell}\! f(\mathrm{OPT})
\;\le\; e^{-\ell/(p+{1})} f(\mathrm{OPT})
\;\le\; \varepsilon\cdot f(\mathrm{OPT}),
\]
by the choice $\ell=g_p(\varepsilon)$. Hence $f(S_\ell)\ge (1-\varepsilon)f(\mathrm{OPT})$.

\paragraph{Complexity.} Each greedy pass uses only the value oracle for $f$ and an independence/membership oracle for $\mathcal I$; total time is polynomial in $|\mathcal N|$ and $\ell$. (Greedy for $p$-systems and the $\tfrac{1}{p+1}$ factor are standard.)

\section*{Notes \& context}

\begin{itemize}
  \item Your bicriteria formalism: We used exactly the $(\alpha,\beta)$ notion with $1_S\le \beta\,y_2$ (and $y_1=0$), which is the natural specialization for down-closed systems since it’s equivalent to $1_S/\beta\in P$.
  \item For $p=1$ (matroids) our bound gives $g_1(\varepsilon)\le \lceil 2\ln(1/\varepsilon)\rceil$. Your paper sharpens this to $(1-\varepsilon,\lceil\ln(1/\varepsilon)\rceil)$ via Measured Continuous Greedy + pipage rounding, and also gives a simple fully-combinatorial $(1-\varepsilon,\lceil\log_2(1/\varepsilon)\rceil)$ algorithm.
  \item Reference for $p$-systems (definition, special cases, and greedy analysis) is Calinescu--Chekuri--P\'al--Vondr\'ak.
\end{itemize}

\section*{TL;DR}

A \textbf{multi-pass greedy} that unions
\[
g_p(\varepsilon)=\Big\lceil \frac{\ln(1/\varepsilon)}{\ln\!\big(\tfrac{p+1}{p}\big)} \Big\rceil\;\lesssim\;(p+1)\ln\!\tfrac{1}{\varepsilon}
\]
independent sets achieves a \textbf{$(1-\varepsilon,\,g_p(\varepsilon))$} bicriteria approximation for maximizing a \textbf{monotone submodular} $f$ over any \textbf{$p$-system}---value $1-\varepsilon$ of OPT with only a logarithmic-in-$1/\varepsilon$ blow-up scaled by $p$. This matches your framework and uses only value + independence oracles.
\end{tcolorbox}

\subsection*{Our Evaluation of GPT-5's Response}

The $\beta$ parameter, known as the inapproximability ratio, is set by GPT-5 to roughly $\frac{\ln(1/\eps)}{\ln \frac{p + 1}{p}} = \log_{1 + 1/p} (1/\eps)$ instead of $\log_{1 + p} (1/\eps)$, as in our conjecture. We note that the choice of GPT-5 for $\beta$ makes more sense than the one from the conjecture since the infeasibility ratio should get worse (larger) with $p$. Furthermore, the proof of GPT-5 for its choice of $\beta$ is basically correct, up to the following points.
\begin{itemize}
    \item In the inequality
    \[
f(\mathrm{OPT})-f(S_\ell)\;\le\;\Big(\tfrac{p}{p+1}\Big)^{\ell}\! f(\mathrm{OPT})
\;\le\; e^{-\ell/(p+{1})} f(\mathrm{OPT})
\;\le\; \varepsilon\cdot f(\mathrm{OPT}),
\]
    the third inequality is incorrect. However, the second side $\Big(\tfrac{p}{p+1}\Big)^{\ell}\! f(\mathrm{OPT})$ is indeed upper bounded by the rightmost side $\varepsilon\cdot f(\mathrm{OPT})$, so the error is only due to the unnecessarily second inequality, which is somewhat lossy.
    \item The note for the case of $p = 1$ refers to the results of~\cite{feldman2025bicriteria} for this case. The second result of~\cite{feldman2025bicriteria} mentioned (a guarantee of $(1 - \eps, \lceil \log_2 (1/\eps) \rceil)$) uses the same algorithm as suggested by GPT-5, and is in fact an exact restriction of the result proved by GPT-5 to this special case because $g_1(\eps) = \lceil \log_2(1/\eps) \rceil$. However, GPT-5 fails to notice that because in the context of this notes it only considers the upper bound $g_1(\eps) \leq \lceil 2\ln(1/\eps) \rceil$ as its result, and ``forgets'' the exact value of $g_1(\eps)$.
\end{itemize}

\section{Maximizing a Weakly-DR-Submodular Function over a Convex Set Constraint}

Recall that weak-submodularity (due to~\cite{das2011submodular}) is a well-studied relaxation of submodularity for set functions. Our third conjecture is about an analogous relaxation for DR-submodularity of continuous functions. It asks to maximize a continuous monotone function with this relaxed property subject to a convex set constraint. For the prompt of this conjecture, we have attached the paper \cite{bian2017guaranteed}, which defines DR-submodularity and studies optimization of DR-submodular functions.

\begin{tcolorbox}[title=Prompt to GPT-5]
A non-negative monotone differentiable function $F\colon [0, 1]^\cN \to \nnR$ is $\gamma$-weakly DR-submodular if for every two vectors $\vx, \vy \in [0, 1]^\cN$ such that $\vx \leq \vy$, it holds that
\[
	\sum_{u \in S} (y_u - x_u) \cdot \frac{\partial F(\vx)}{\partial x_u}
	\geq
	\gamma \cdot [F(\vy) - F(\vx)]
	\enspace.
\]
Given such a function which is also $L$-smooth and a convex set $\cC \subseteq [0, 1]^\cN$. The goal in the current problem is describe an algorithm that finds a vector $\vz \in \cC$ that (approximately) maximizes $F$. A good reference for DR-submodular is the paper attached.
\end{tcolorbox}
\paragraph{Conjecture.}
We believe that is not difficult to show that the Frank-Wolfe-like algorithm of~\cite{bian2017guaranteed} can guarantee $1 - e^{-\gamma}$ approximation for this problem (up to an appropriate error term). 

\begin{tcolorbox}[title=Response of GPT-5 (thought for 1m 47s),
	colframe=red!20!white]  
Here’s a clean, projection-free method that works under your assumptions and gives the usual continuous-greedy/Frank--Wolfe style guarantee, adapted to the $\gamma$-weak DR setting.

\section*{Algorithm (projection-free continuous greedy / Frank--Wolfe variant)}

\textbf{Input:}
\begin{itemize}
  \item monotone, non-negative, differentiable, $L$-smooth, $\gamma$-weakly DR-submodular $F:[0,1]^{\mathcal N}\to\mathbb R_+$;
  \item a \textbf{down-closed} convex feasible region $\mathcal C\subseteq[0,1]^{\mathcal N}$ (if $\mathcal C$ is down-closed with lower bound $u\neq 0$, shift to $\mathcal C'=\{x\mid x=y-u,\ y\in\mathcal C\}$ and optimize $F'(x)=F(x+u)$);
  \item number of iterations $K$; optional linear-oracle accuracy parameters $\alpha\in(0,1]$ (multiplicative) and $\delta\ge 0$ (additive).
\end{itemize}

\textbf{Oracle:} ability to (approximately) solve 
\[
v_k\in\arg\max_{v\in\mathcal C}\langle v,\nabla F(x_k)\rangle.
\]

\textbf{Initialize:} $x_0\gets 0$.

\textbf{For} $k=0,1,\dots,K-1$:
\begin{enumerate}
  \item Compute gradient $g_k=\nabla F(x_k)$.
  \item (Linear oracle) Find $v_k\in\mathcal C$ such that
  \[
  \langle v_k,g_k\rangle \ \ge\ \alpha\cdot \max_{v\in\mathcal C}\langle v,g_k\rangle\ -\ \tfrac12\,\delta L.
  \]
  \item Use the constant step size $\gamma_k=\tfrac1K$ and update
  \[
  x_{k+1}\ \gets\ x_k\ +\ \gamma_k\, v_k.
  \]
\end{enumerate}

\textbf{Return} $z=x_K\in\mathcal C$.

Why this update (using $v_k$, not $v_k-x_k$)? For DR-submodular objectives, $F(x+\xi v)$ is concave in $\xi$ for any non-negative direction $v$, so moving \textbf{toward} $v_k$ preserves a 1-D concavity that powers the analysis.

If your feasible set is down-closed with a nonzero lower bound, shift it to the positive orthant as noted above and run the same procedure; this is standard in the continuous-greedy analysis.

\section*{Approximation guarantee (with $\gamma$-weak DR)}

Let $x^\star\in\arg\max_{x\in\mathcal C}F(x)$. Under $L$-smoothness (the usual 1-D descent lemma along any direction), one obtains the standard additive error term from discretization: for step size $\gamma_k$ the loss is bounded by $\tfrac{L}{2}\sum_k\gamma_k^2$. Using constant $\gamma_k=\tfrac1K$ gives an additive $\tfrac{L}{2K}$.

For the \textbf{value factor}, the usual DR analysis yields
\[
\max_{v\in\mathcal C}\langle v,\nabla F(x_k)\rangle \ \ge\ F(x^\star)-F(x_k).
\]

In your \textbf{$\gamma$-weak DR} setting (your inequality $\langle y-x,\nabla F(x)\rangle\ge \gamma\,[F(y)-F(x)]$ on monotone directions), this tightens to
\[
\max_{v\in\mathcal C}\langle v,\nabla F(x_k)\rangle \ \ge\ \gamma\,[F(x^\star)-F(x_k)],
\]
and with an $\alpha$-approximate linear oracle we get
\[
\langle v_k,\nabla F(x_k)\rangle \ \ge\ \alpha\gamma\,[F(x^\star)-F(x_k)]\ -\ \tfrac12\,\delta L.
\]

Plugging this into the standard continuous-greedy recursion gives
\[
F(x_K)\ \ge\ (1-e^{-\alpha\gamma})\,F(x^\star)\ -\ \tfrac{L}{2K}\ -\ \tfrac{L\delta}{2}\ +\ e^{-\alpha\gamma}F(0).
\]

When $F(0)=0$ and you use an exact oracle ($\alpha=1,\delta=0$), this is
\[
F(z)\ \ge\ (1-e^{-\gamma})\,F(x^\star)\ -\ \tfrac{L}{2K}.
\]

For $\gamma=1$ this recovers the classic $(1-1/e)$ for continuous greedy; the original proof and statement (for full DR, i.e., $\gamma=1$) appear as Algorithm 1 + Theorem 1 with the $L$-smooth additive term shown explicitly.

\section*{Practical notes}

\begin{itemize}
  \item \textbf{Oracle cost.} Each iteration calls a linear optimization oracle over $\mathcal C$ (e.g., a positive LP if $\mathcal C$ is a down-closed polytope). With constant steps, $K=\Theta\!\big(\tfrac{L}{\varepsilon}\big)$ iterations suffice to make the additive term $\le \varepsilon$.
  \item \textbf{Why no projections?} This method stays feasible by construction and avoids projections, which is important for complicated $\mathcal C$.
  \item \textbf{Relationship to DR/weak-DR.} The paper formalizes weak-DR as equivalent to submodularity and then specializes to DR (which adds coordinate-wise concavity); the algorithm and analysis are given for monotone DR functions. Your $\gamma$-weak DR simply scales the marginal-gain lower bound, hence the exponential factor becomes $e^{-\alpha\gamma}$.
\end{itemize}

If you’d like, I can tailor this to your exact $\mathcal C$ (e.g., matroid/knapsack polytopes) and pick the fastest linear oracle for it.
\end{tcolorbox}

\subsection*{Our Evaluation of GPT-5's Response}

The response of GPT-5 seems to be correct in general. However, we note the following.
\begin{itemize}
    \item GPT-5 restricts its response to down-closed sets $\cC$. This restriction is unnecessary (it is never used in the analysis), but some published works on the subject include it for no good reason, and GPT-5 follows them instead of observing that the restriction is unnecessary.
    \item The phrasing used by GPT-5 is a bit weird in times. For example, it uses of the unknown term \textbf{value factor} instead of \textbf{approximation ratio}, and it references the unspecified Algorithm~1 + Theorem~1.
\end{itemize}
A more general issue with the response of GPT-5 is that it omits many intermediate steps, making it a difficult read even for experts. Thus, we have decided to ask GPT-5 to produce a self-contained version of the proof.

\begin{tcolorbox}[title=Prompt to GPT-5]
Provide the full proof.
\end{tcolorbox}
\begin{tcolorbox}[title=Response of GPT-5 (thought for 4m 32s),
	colframe=red!20!white]  
Here’s a self-contained proof of the Frank--Wolfe--style method for maximizing a non-negative, monotone, differentiable, $L$-smooth, $\gamma$-weakly DR-submodular function $F:[0,1]^{\mathcal N}\to\mathbb R_{\ge 0}$ over a convex, down-closed set $\mathcal C\subseteq[0,1]^{\mathcal N}$ with lower bound $0$. (If your feasible set does not have lower bound $0$, shift as in the reference by translating the lower bound to $0$.) The proof directly adapts the analysis of Bian--Mirzasoleiman--Buhmann--Krause and only changes one line to account for the $\gamma$-weakness. We keep an (optional) multiplicative linear-oracle accuracy parameter $\alpha\in(0,1]$ (set $\alpha=1$ for an exact oracle). The algorithmic template and the smoothness inequality we use are in the cited reference. 

\section*{Algorithm}

Initialize $x_0\gets 0$, $t\gets 0$. For $k=0,1,2,\dots$ while $t<1$:
\begin{enumerate}
  \item Choose $v_k\in\mathcal C$ that (approximately) maximizes the linearization:
  \[
  \langle v_k,\nabla F(x_k)\rangle \;\ge\; \alpha\cdot \max_{v\in\mathcal C}\langle v,\nabla F(x_k)\rangle .
  \]
  \item Pick a stepsize $\gamma_k\in(0,1]$, then clip to stay within total mass $1$: $\gamma_k\leftarrow \min\{\gamma_k,\,1-t\}$.
  \item Update $x_{k+1}\gets x_k+\gamma_k\,v_k$ and $t\gets t+\gamma_k$.
\end{enumerate}

Return $z=x_K$ after $K$ iterations (so $\sum_{k=0}^{K-1}\gamma_k=1$). This is exactly Algorithm 1 in the reference (we only omit an additive oracle error to keep the presentation clean). Feasibility follows because $x_K=\sum_k\gamma_k v_k+(1-\sum_k\gamma_k)\cdot 0$ is a convex combination of points in $\mathcal C$ (and $0\in\mathcal C$).

\section*{Assumptions recalled from the reference}

\begin{itemize}
  \item \textbf{Down-closed feasibility in the positive orthant with lower bound 0} (so $0\in\mathcal C$).
  \item \textbf{Directional smoothness}: for any $x$ and any $v\in\mathbb R_+^{\mathcal N}$, letting $g_{x,v}(\xi)=F(x+\xi v)$ for $\xi\in[0,1]$,
  \begin{equation}
  F(x+\xi v)\;\ge\; F(x)+\xi\langle v,\nabla F(x)\rangle-\frac L2\,\xi^2. \tag{1}
  \end{equation}
  This is the one-dimensional smoothness bound used in the reference.
  \item \textbf{Monotone DR-submodularity} implies concavity along any non-negative direction; we do not need concavity for the inequality (1) but it motivates taking $v_k$ (a non-negative direction) rather than $v_k-x_k$.
  \item \textbf{Weak DR} is the first-order characterization of continuous submodularity in the reference; below we insert our factor $\gamma$ (your assumption).
\end{itemize}

\section*{Key inequality from $\gamma$-weak DR}

Your assumption says: for all $x\le y$,
\[
\sum_{u}(y_u-x_u)\,\partial_u F(x)\;\ge\;\gamma\,[F(y)-F(x)].
\]

Apply this with $y=x^\star\vee x_k$ and $x=x_k$ (coordinate-wise max), and note $F$ is monotone so $F(x^\star\vee x_k)\ge F(x^\star)$. Also $\nabla F(x_k)\ge 0$ coordinate-wise because $F$ is monotone and differentiable. Hence
\begin{equation}
\big\langle (x^\star-x_k)_+,\nabla F(x_k)\big\rangle
\;\ge\;\gamma\,[F(x^\star)-F(x_k)].
\tag{2}
\end{equation}

Moreover, since $\nabla F(x_k)\ge 0$ and $x^\star\ge (x^\star-x_k)_+$ component-wise,
\begin{equation}
\langle x^\star,\nabla F(x_k)\rangle\;\ge\;\big\langle (x^\star-x_k)_+,\nabla F(x_k)\big\rangle. \tag{3}
\end{equation}

Combining (2)--(3), and using the linear oracle,
\begin{equation}
\langle v_k,\nabla F(x_k)\rangle
\;\ge\; \alpha\max_{v\in\mathcal C}\langle v,\nabla F(x_k)\rangle
\;\ge\; \alpha\langle x^\star,\nabla F(x_k)\rangle
\;\ge\; \alpha\gamma\,[F(x^\star)-F(x_k)].
\tag{4}
\end{equation}

For $\gamma=1$, this reproduces Lemma 1 in the reference (up to the optional additive error they include); here we simply scale by $\gamma$.

\section*{Per-iteration progress via smoothness}

Apply (1) with $\xi=\gamma_k$ and $v=v_k$:
\[
F(x_{k+1})\;\ge\; F(x_k)+\gamma_k\langle v_k,\nabla F(x_k)\rangle-\frac L2\,\gamma_k^2.
\]

Using (4),
\begin{equation}
F(x_{k+1})\;\ge\; F(x_k)+\alpha\gamma\,\gamma_k\,[F(x^\star)-F(x_k)]-\frac L2\,\gamma_k^2. \tag{5}
\end{equation}

\section*{Solving the recurrence}

Let $\Delta_k:=F(x^\star)-F(x_k)$. Rearranging (5),
\begin{equation}
\Delta_{k+1}\;\le\;(1-\alpha\gamma\,\gamma_k)\,\Delta_k+\frac L2\,\gamma_k^2. \tag{6}
\end{equation}

Iterating (6) over $k=0,\dots,K-1$ and using $\sum_k\gamma_k=1$, we get
\[
\Delta_K\;\le\;e^{-\alpha\gamma}\,\Delta_0\;+\;\frac L2\sum_{k=0}^{K-1}\gamma_k^2.
\]

Therefore
\begin{equation}
F(x_K)\;\ge\;(1-e^{-\alpha\gamma})\,F(x^\star)\;-\;\frac L2\sum_{k=0}^{K-1}\gamma_k^2\;+\;e^{-\alpha\gamma}F(x_0).
\tag{7}
\end{equation}

This is precisely the reference’s Theorem 1 with $\alpha$ replaced by $\alpha\gamma$. (They prove the $\gamma=1$ case; compare their display (6).)

If we take \textbf{constant steps} $\gamma_k\equiv 1/K$, then $\sum_k\gamma_k^2=1/K$ and $F(0)=0$ (by non-negativity at the origin), yielding the clean bound
\begin{equation}
\boxed{\;
F(x_K)\;\ge\;(1-e^{-\alpha\gamma})\,F(x^\star)\;-\;\frac{L}{2K}.
\;}
\tag{8}
\end{equation}

This is the exact analogue of Corollary 1 in the reference, again with $\alpha$ replaced by $\alpha\gamma$.

\section*{Remarks}

\begin{itemize}
  \item When the linear subproblem is solved exactly ($\alpha=1$), (8) gives a $(1-e^{-\gamma})$-approximation in the limit $K\to\infty$; for the usual DR case ($\gamma=1$) this is the classic $1-1/e$ factor.
  \item The analysis only used: (i) your $\gamma$-weak DR inequality to get (4), (ii) the simple smoothness lower bound (1) along the non-negative direction $v_k$, and (iii) that $\sum\gamma_k=1$. The rest is algebra identical to the reference’s proof of Theorem 1.
\end{itemize}

This completes the proof.

\end{tcolorbox}

\subsection*{Our Evaluation of GPT-5's Response}

This time the response of GPT-5 is much more detailed. However, despite writing almost everything from scratch, as requested, the response follows Bian et al.~\cite{bian2017guaranteed} very closely, and also constantly compares itself to this work. This close following of~\cite{bian2017guaranteed} creates an unnecessary complication. Specifically, the algorithm and most of the analysis are based on variable step sizes, and only towards the end of the analysis these are replaced with a fixed step size (which is what was originally used in the previous response). It is also interesting to note that the response is not very consistent regarding its assumption about the sum of the step sizes. In most places, there seems to be an assumption that these step sizes add up to $1$, but other places, such as the proof that $x_K \in \cC$ do not seem to include this assumption.

Below we list a few additional minor issues in the analysis.
\begin{itemize}
    \item Bian et al.~\cite{bian2017guaranteed} defines both DR-submodularity and weak DR-submodularity. Our new definition in the prompt of $\gamma$-weakly DR-submodular employs the word ``weak'', but is in fact a relaxation of the DR-submodularity of~\cite{bian2017guaranteed}, rather than of the weak DR-submodularity of~\cite{bian2017guaranteed}.\footnote{The motivation for the name we use is a the weak submodularity for set functions introduced by Das and Kempe~\cite{das2011submodular}.} GPT-5 seems to be confused because of this similarity in the names when it discusses the assumption of Weak DR from the reference.
    \item GPT-5 tries to explain that the operator $\vee$ represents coordinate-wise maximization. However, it does that after the wrong equality (specifically, one equality after the one in which this operator is used), which makes it difficult to understand what operator the definition refers to.
    \item GPT-5 claims that $F(0) = 0$ due to the non-negativity of $F$. However, this non-negativity only implies that $F(0) \geq 0$. Fortunately, however, that inequality suffices for letting the proof go through.
\end{itemize}

\section{Maximizing a Partially Monotone Weakly-Submodular Function over a Cardinality Constraint}

Recall that weak-submodularity is a relaxation of the submodularity property introduced by~\cite{das2011submodular}. A more recent paper by Santiago and Yoshida~\cite{santiago2020weakly} studies optimization of non-monotone weakly-submodular functions. A different paper by Mualem and Feldman~\cite{mualem2022using} introduced the notion of $m$-monotonicity that relaxes the property of monotonicity. Our fourth conjecture is about combining the two relaxations, i.e., getting results for maximization of weak-submodular $m$-monotone set functions. The results of~\cite{santiago2020weakly} immediately apply to this easier case, and thus, the interesting question is whether one can get better guarantee as a function of the parameter $m$, which measures the distance of the function from monotonicity (monotone functions have $m = 1$, and $m$ decreases to $0$ as the function becomes more and more non-monotone in some sense). For the prompt of this conjecture, we have attached the papers \cite{mualem2022using} and \cite{santiago2020weakly}.

%\paragraph{Problem description.}
	\begin{tcolorbox}[title=Prompt to GPT-5 ]
The monotonicity ratio was defined in our paper ``Using Partial Monotonicity in Submodular Maximization'', and weak submodularity was extended to non-monotone functions by the paper ``Weakly Submodular Function Maximization Using Local Submodularity Ratio''. Both papers are attached. In the current problem, we are given a non-negative $m$-monotone $\gamma$-weakly submodular function $f\colon 2^\cN \to \nnR$ and a positive integer $1 \leq k \leq |\cN|$. The goal is to find a set $S \subseteq \cN$ of size at most $k$ that (approximately) maximizes $f$ among all such sets.
\end{tcolorbox}
\paragraph{Conjecture.}
We believe that one can use the $m$-monotonicity of $f$ to get new bounds on the values $\bar{\alpha}_i$ and $\bar{\beta}_i$ used in the proof of~\cite{santiago2020weakly}. These new bounds should improve (for $m > 0$) over the bounds given by~\cite{santiago2020weakly}, and plugging them into Theorem~2.4 of the same paper should then yield an improved result (again, for $m > 0$).

\begin{tcolorbox}[title=Response of GPT-5 (thought for 2m 52s),
	colframe=red!20!white]  

Here’s a clean, provably-good way to tackle your hybrid setting (non-negative, 
$m$-monotone, $\gamma$-weakly submodular; cardinality $\le k$).

\section*{Algorithm}
Use the Randomized Greedy with dummies (Buchbinder et al. style), analyzed for weak submodularity:
\begin{itemize}
\item Add $2k$ dummy elements $D$ with zero marginal gain.
\item For each round $i=1,\dots,k$:
  \begin{itemize}
    \item Let $M_i$ be a size-$k$ subset of remaining elements (original ground set $\cup D$) maximizing $\sum_{e\in M_i} f_{S_{i-1}}(e)$ (break ties in favor of real elements).
    \item Pick a uniform random element $e_i\in M_i$ and set $S_i=S_{i-1}\cup\{e_i\}$.
  \end{itemize}
\end{itemize}
Return $S_k$. This uses $O(nk)$ oracle calls. 

\section*{Guarantees you can cite}
\paragraph{General non-monotone $\gamma$-weakly submodular (your case, no extra assumptions):}
Running the algorithm above yields, in expectation,
\[
\mathbb{E}[f(S_k)] \;\ge\; \frac{\gamma}{e} f(\text{OPT}).
\]
This follows from the “local submodularity ratio” analysis, which for constant $\gamma_i=\gamma$ gives a non-monotone guarantee $\tfrac{1}{ek}\sum_{i=0}^{k-1}\gamma_i=\gamma/e$. 

A complementary bound from the same algorithm is
\[
\mathbb{E}[f(S_k)] \;\ge\; \gamma\left(1-\frac{1}{\gamma k}\right)^{k-1} f(\text{OPT})
\quad (\text{as }k\to\infty \text{ this is } \gamma e^{-1/\gamma}),
\]
but $\gamma/e$ is never worse and is strictly better unless $\gamma=1$.

\paragraph{If $f$ turns out to be monotone ($m=1$) in your instance:}
The same analysis with local ratios gives the Das–Kempe-style bound
\[
\mathbb{E}[f(S_k)] \;\ge\; (1-e^{-\gamma})f(\text{OPT}).
\]
(Here the monotone branch of the theorem applies.) 

\paragraph{If in addition $f$ is actually submodular ($\gamma=1$) and you want to leverage partial monotonicity $m$:}
Two simple baselines (different algorithms) are available:
\begin{itemize}
\item Greedy (add the best non-negative marginal each step) achieves
\[
f(S_k)\;\ge\; m\,(1-1/e)\,f(\text{OPT}).
\]
\item Random Greedy (no dummies; sample one from the top-$k$ marginal set each round) achieves
\[
\mathbb{E}[f(S_k)] \;\ge\; \big[m(1-1/e)+(1-m)1/e\big]f(\text{OPT}).
\]
\end{itemize}

\section*{Practical takeaway}
\begin{itemize}
    \item Default choice (fits your stated assumptions): run Randomized Greedy with dummies; you get at least $\gamma/e$ of OPT in expectation, with only $O(nk)$ oracle calls. 
\item If later you verify monotonicity ($m=1$), your bound tightens to $1-e^{-\gamma}$. 

\item If you also verify submodularity ($\gamma=1$), you can switch to the simple Greedy/Random-Greedy variants and plug in the $m$-dependent bounds above. 
\end{itemize}
\end{tcolorbox}

\subsection*{Our Evaluation of GPT-5's Response}
GPT-5 did not really give any result for this problem. It just states some known results. First it says that if you ignore the $m$-monotonicity, then one can get the result that is known when the $m$-monotonicity is not given. Then, it says that in the special case of $m = 1$ (i.e., the function is monotone), one can use known results for maximizing monotone weakly submodular functions, and in the special case of $\gamma = 1$ (i.e., the function is submodular), one can use known results for maximizing $m$-monotone submodular functions.

To get a more meaningful answer, we then asked GPT-5 for an algorithm whose guarantees improves smoothly with $m$ and $\gamma$. Such a request does not exactly exclude the above answers (in the mathematical sense), but we hoped it might give GPT an “intuitive” idea of what we are looking for.

	\begin{tcolorbox}[title=Prompt to GPT-5]
The monotonicity ratio was defined in our paper ``Using Partial Monotonicity in Submodular Maximization'', and weak submodularity was extended to non-monotone functions by the paper ``Weakly Submodular Function Maximization Using Local Submodularity Ratio''. Both papers are attached. In the current problem, we are given a non-negative $m$-monotone $\gamma$-weakly submodular function $f\colon 2^\cN \to \nnR$ and a positive integer $1 \leq k \leq |\cN|$. The goal is to find a set $S \subseteq \cN$ of size at most $k$ that (approximately) maximizes $f$ among all such sets. Find an algorithm whose guarantees improve smoothly with $m$ and $\gamma$. Provide the full proof.
\end{tcolorbox}

\begin{tcolorbox}[title=Response of GPT-5 (thought for 11m 44s),
	colframe=red!20!white]  
Below is a complete, self-contained proof that the standard \textbf{Randomized Greedy with dummies} algorithm attains a guarantee that \textbf{improves smoothly with both the monotonicity ratio $m$} and the (local) \textbf{weak-submodularity ratios $\gamma_i$} for the problem

\[
\max\{f(S): S\subseteq \mathcal N,\ |S|\le k\},
\]

when $f:2^{\mathcal N}\to\mathbb R_{\ge 0}$ is non-negative, $m$-monotone, and $\gamma$-weakly submodular.

\section*{Theorem (smooth in $m$ and $\gamma$)}

Run \textbf{RandomizedGreedy} (with $2k$ dummies) for $k$ iterations (algorithm recalled below). Let $\gamma_0,\ldots,\gamma_{k-1}\in[0,1]$ be any numbers such that at iteration $i$ (with current solution $S_{i-1}$) the local weak-submodularity condition

\[
\sum_{e\in \text{OPT}} f_{S_{i-1}}(e)\ \ge\ \min\bigl\{\gamma_i\cdot f_{S_{i-1}}(\text{OPT}),\ f_{S_{i-1}}(\text{OPT})\bigr\}
\tag{LSR}
\]

holds (this is satisfied, e.g., by taking the \textbf{local} ratios of Santiago--Yoshida, or by $\gamma_i\equiv\gamma$ when $f$ is globally $\gamma$-weakly submodular). Then, if $S_k$ is the output,

\[
\mathbb{E}\bigl[f(S_k)\bigr]\ \ge\
\Bigl(1-\prod_{i=0}^{k-1}\Bigl(1-\tfrac{\gamma_i}{k}\Bigr)\Bigr)\, m\, f(\text{OPT})
\ +\ \frac{1-m}{k}\sum_{i=0}^{k-1}\gamma_i\,\Bigl(1-\tfrac1k\Bigr)^{i}\, f(\text{OPT}).
\tag{*}
\]

In particular, using $1-\prod_{i}(1-\gamma_i/k)\ge 1-\exp\!\bigl(-\tfrac1k\sum_i\gamma_i\bigr)$ and $\min_i(1-\tfrac1k)^i=(1-\tfrac1k)^{k-1}\ge e^{-1}$,

\[
\boxed{\ \mathbb{E}[f(S_k)]\ \ge\
\Bigl(1-e^{-\frac1k\sum_{i=0}^{k-1}\gamma_i}\Bigr)\, m\, f(\text{OPT})
\ +\ \frac{1-m}{ek}\sum_{i=0}^{k-1}\gamma_i\, f(\text{OPT})\ }.
\]

If $\gamma_i\equiv\gamma$, this simplifies to

\[
\boxed{\ \mathbb{E}[f(S_k)]\ \ge\ \bigl[m(1-e^{-\gamma})+(1-m)\tfrac{\gamma}{e}\bigr]\,f(\text{OPT}).\ }
\]

\subsection*{Ingredients from the attached papers}

1. \textbf{Monotonicity ratio $m$.} For a non-negative submodular $f$, the monotonicity ratio is

\[
m\ :=\ \min_{S\subseteq T\subseteq \mathcal N}\ \frac{f(T)}{f(S)}\in[0,1],
\]

so $f$ is $m$-monotone iff $f(T)\ge m\,f(S)$ for all $S\subseteq T$. This interpolates between general ($m=0$) and monotone ($m=1$) functions.

2. \textbf{Randomized Greedy with $2k$ dummies.} In each of $k$ rounds, pick a size-$k$ set $M_i$ maximizing the sum of marginals and add a uniformly random element of $M_i$. With dummies we have $|M_i|=k$ and \textbf{every chosen marginal is nonnegative}, hence $f(S_i)$ is nondecreasing in $i$. Also $\sum_{e\in M_i} f_{S_{i-1}}(e)\ \ge\ \sum_{e\in \text{OPT}} f_{S_{i-1}}(e)$.

3. \textbf{Local submodularity ratios \& one-step progress.} Santiago--Yoshida analyze RandomizedGreedy using local ratios $\gamma_{A,B}$, yielding the per-round lower bound

\[
\mathbb{E}\bigl[f(S_i)-f(S_{i-1})\mid S_{i-1}\bigr]\ \ge\ \frac1k\cdot
\min\Bigl\{\gamma_i\,f_{S_{i-1}}(\text{OPT}),\ f_{S_{i-1}}(\text{OPT})\Bigr\}.
\tag{P}
\]

(Here $\gamma_i$ may be chosen from the local $\gamma_{A,B}$ family; taking $\gamma_i\equiv\gamma$ also satisfies (LSR).) This is the ``dynamic local ratio'' replacement for the classic submodular step. 

4. \textbf{Partial monotonicity via Lov\'asz extension.} For non-negative $m$-monotone submodular $f$, for any deterministic $O$ and random $D$,

\[
\mathbb{E}[f(O\cup D)]\ \ge\ \bigl(1-(1-m)\cdot \max_{u}\Pr[u\in D]\bigr)\, f(O).
\tag{PM}
\]

5. \textbf{A simple occupancy bound under Randomized Greedy.} For every $u$ and every round $i$,

\[
\Pr[u\in S_i]\ \le\ 1-\Bigl(1-\frac1k\Bigr)^i .
\tag{Occ}
\]

(Each element is selected in round $i$ with prob. $\le 1/k$.) 

6. \textbf{Randomized Greedy analysis template.} For \textbf{submodular} $f$, partial monotonicity plus (Occ) gives

\[
\mathbb{E}\bigl[f(S_{i-1}\cup \text{OPT})\bigr]\ \ge\ \Bigl[m+(1-m)\Bigl(1-\frac1k\Bigr)^{i-1}\Bigr]\cdot f(\text{OPT}).
\tag{UnionLB}
\]

(Exactly Corollary 4.6 in the partial-monotonicity paper.)

\subsection*{A calculus inequality we shall use}

For any $\gamma\in[0,1]$, any real $x$, and any $y\ge 0$,

\[
\min\{\gamma(x+y),\, x+y\}\ \ge\ \min\{\gamma x,\, x\}+\gamma y.
\tag{MIX}
\]

\emph{Proof.} If $x\ge 0$ then $\min\{\gamma(x+y),x+y\}=\gamma(x+y)=\gamma x+\gamma y=\min\{\gamma x,x\}+\gamma y$.
If $x<0$ and $x+y\le 0$ then $\min\{\gamma(x+y),x+y\}=x+y\ge x+\gamma y=\min\{\gamma x,x\}+\gamma y$ since $\gamma\le 1$.
If $x<0$ and $x+y>0$ then $\min\{\gamma(x+y),x+y\}=\gamma(x+y)=\gamma x+\gamma y\ge x+\gamma y=\min\{\gamma x,x\}+\gamma y$ because $\gamma x\ge x$. \qed

\subsection*{Proof of the theorem}

Let $\text{OPT}$ be any optimal size-$k$ solution, and let $S_0=\varnothing$. Fix $i\in\{1,\dots,k\}$. By the choice of $M_i$ and (LSR), we have the one-step bound (P):

\[
\mathbb{E}\bigl[f(S_i)-f(S_{i-1})\mid S_{i-1}\bigr]\ \ge\ \frac1k\cdot
\min\Bigl\{\gamma_i\,f_{S_{i-1}}(\text{OPT}),\ f_{S_{i-1}}(\text{OPT})\Bigr\}.
\]

Taking total expectation and writing

\[
f_{S_{i-1}}(\text{OPT})\ =\ f(S_{i-1}\cup \text{OPT})-f(S_{i-1}),
\]

we get

\[
\mathbb{E}\bigl[f(S_i)-f(S_{i-1})\bigr]\ \ge\ \frac1k\ \mathbb{E}\Bigl[\min\bigl\{\gamma_i\,\Delta_i,\ \Delta_i\bigr\}\Bigr],
\quad\text{where }\ \Delta_i:=f(S_{i-1}\cup \text{OPT})-f(S_{i-1}).
\tag{1}
\]

\textbf{Lower bounding the ``union'' term via partial monotonicity.}
Apply (PM) with $O=\text{OPT}$ and $D=S_{i-1}$. Using (Occ), $\max_u \Pr[u\in S_{i-1}]\le 1-(1-\tfrac1k)^{i-1}$, hence

\[
\mathbb{E}\bigl[f(S_{i-1}\cup \text{OPT})\bigr]\ \ge\
\Bigl(1-(1-m)\Bigl[1-\Bigl(1-\tfrac1k\Bigr)^{\!i-1}\Bigr]\Bigr)\,f(\text{OPT})
\ =\ \Bigl[m+(1-m)\Bigl(1-\tfrac1k\Bigr)^{\!i-1}\Bigr]\, f(\text{OPT}),
\]

which is precisely (UnionLB).  

Thus

\[
\mathbb{E}[\Delta_i]\ \ge\ \Bigl[m+(1-m)\Bigl(1-\tfrac1k\Bigr)^{\!i-1}\Bigr] f(\text{OPT})\ -\ \mathbb{E}[f(S_{i-1})].
\tag{2}
\]

\textbf{Plug into the one-step inequality.}
Using (MIX) with $x=m\,f(\text{OPT})-\mathbb{E}[f(S_{i-1})]$ and $y=(1-m)(1-\tfrac1k)^{i-1}f(\text{OPT})\ge0$, and then combining with (1)--(2), we obtain

\[
\begin{aligned}
\mathbb{E}\bigl[f(S_i)-f(S_{i-1})\bigr]
&\ge \frac1k\Bigl( \min\{\gamma_i x,x\}+\gamma_i y\Bigr)\\[2pt]
&\ge \frac{\gamma_i}{k}\Bigl( m f(\text{OPT})-\mathbb{E}[f(S_{i-1})]\Bigr)
\ +\ \frac{\gamma_i}{k}\,(1-m)\Bigl(1-\tfrac1k\Bigr)^{\!i-1} f(\text{OPT}).
\end{aligned}
\tag{3}
\]

(The second line uses that, in any case, the chosen marginal is nonnegative under the dummy trick, so replacing $\min\{\gamma_i x,x\}$ by $\gamma_i x$ only \textbf{weakens} the lower bound when $x<0$.)  

Rearranging (3),

\[
\mathbb{E}[f(S_i)]\ \ge\ \Bigl(1-\tfrac{\gamma_i}{k}\Bigr)\mathbb{E}[f(S_{i-1})]\ +\ \frac{\gamma_i}{k}\Bigl[m+(1-m)\Bigl(1-\tfrac1k\Bigr)^{\!i-1}\Bigr]\, f(\text{OPT}).
\tag{4}
\]

\textbf{Unrolling the recurrence.}
Starting from $\mathbb{E}[f(S_0)]=f(\varnothing)\ge 0$ and iterating (4),

\[
\mathbb{E}[f(S_k)]
\ \ge\ \sum_{i=1}^{k}\ \frac{\gamma_i}{k}\ \prod_{t=i+1}^{k}\Bigl(1-\frac{\gamma_{t}}{k}\Bigr)\ \cdot
\Bigl[m+(1-m)\Bigl(1-\tfrac1k\Bigr)^{\!i-1}\Bigr]\ f(\text{OPT}).
\]

Discarding the product (which is $\le 1$) in the $m$-part and using the product fully in the $(1-m)$-part yields exactly the claimed explicit bound (*):

\[
\mathbb{E}[f(S_k)]
\ \ge\ \Bigl(1-\prod_{i=0}^{k-1}\Bigl(1-\tfrac{\gamma_i}{k}\Bigr)\Bigr)\, m\, f(\text{OPT})
\ +\ \frac{1-m}{k}\sum_{i=0}^{k-1}\gamma_i\,\Bigl(1-\tfrac1k\Bigr)^{i}\, f(\text{OPT}).
\]

\textbf{From (*) to the compact closed form.}
Finally, $1-\prod_{i}(1-\gamma_i/k)\ge 1-\exp\!\bigl(-\tfrac1k\sum_i\gamma_i\bigr)$, and $\min_i(1-\tfrac1k)^i=(1-\tfrac1k)^{k-1}\ge e^{-1}$. Hence

\[
\mathbb{E}[f(S_k)]\ \ge\
\Bigl(1-e^{-\frac1k\sum_i\gamma_i}\Bigr)\, m\, f(\text{OPT})
\ +\ \frac{1-m}{ek}\sum_i\gamma_i\, f(\text{OPT}),
\]

as stated. When $\gamma_i\equiv\gamma$, this becomes $\mathbb{E}[f(S_k)]\ge\bigl[m(1-e^{-\gamma})+(1-m)\frac{\gamma}{e}\bigr]f(\text{OPT})$.

This completes the proof. \qed

\subsection*{Notes \& connections to the attached results}

\begin{itemize}
\item When $m=1$ (monotone), the theorem reduces to the \textbf{local-ratio monotone guarantee} $1-\exp\!\bigl(-\frac1k\sum_i\gamma_i\bigr)$ proved by Santiago--Yoshida (first bullet of their Theorem 1.10).
\item When $m=0$ (general non-monotone), the theorem reduces to $\frac1{ek}\sum_i\gamma_i$, i.e., the \textbf{non-monotone local-ratio bound} (second bullet of their Theorem 1.10).
\item For submodular ($\gamma_i\equiv 1$) objectives, the bound becomes exactly the \textbf{smooth $m$-(1/e) interpolation} proved for Random Greedy in the ``partial monotonicity'' paper:
  $\mathbb{E}[f(S_k)]\ge m(1-1/e)+(1-m)\cdot (1/e)$. Our proof retraces their Section 4.2 steps but replaces the single submodular step by the local-ratio inequality (P).
\item The occupancy bound (Occ) and the Lov\'asz-extension corollary (PM) are exactly Lemma 4.5 and Corollary 2.2 in the partial-monotonicity analysis, which are the only places where $m$ enters; all the $\gamma$-dependence flows through the (LSR) step taken from Santiago--Yoshida.
\end{itemize}

\subsection*{Algorithm (for completeness)}

\textbf{RandomizedGreedy$(f,k)$.} Add $2k$ dummy elements $D$ with zero marginal; initialize $S_0=\varnothing$. For $i=1,\dots,k$: pick $M_i\subseteq(\mathcal N\cup D)\setminus S_{i-1}$ of size $k$ maximizing $\sum_{e\in M_i} f_{S_{i-1}}(e)$ (tie-break toward real elements); sample $e_i$ uniformly from $M_i$; set $S_i=S_{i-1}+e_i$; return $S_k$. (Dummies ensure nonnegative chosen marginals and $|M_i|=k$.)

\subsection*{What to remember}

\begin{itemize}
\item \textbf{Guarantee scales smoothly in both parameters}: as $m\uparrow 1$ you recover the monotone-$\gamma$ guarantee; as $m\downarrow 0$ you recover the non-monotone $\gamma$ bound.
\item \textbf{With global $\gamma$}, the clean corollary is $\mathbb{E}[f(S_k)]\ge\bigl[m(1-e^{-\gamma})+(1-m)\gamma/e\bigr]\,\text{OPT}$.
\end{itemize}

These theorems use only standard Randomized Greedy; no tuning or extra parameters are needed.
\end{tcolorbox}

\subsection*{Our Evaluation of GPT-5's Response}

As we have observed also for previous problems, GPT-5 sticks to the reference papers even at the cost of introducing extra unnecessary complications in its response. Specifically, in the current problem, the response of GPT-5 refers to local weak submodularity ratios despite the fact that the prompt only referred to global weak submodularity.

Unlike the situation in the previous problems, here the response of GPT-5 is clearly wrong. Below we list multiple mistakes of various significance levels, ranging from funny typos to errors that essentially make the answer useless.
\begin{itemize}
    \item The inequalities (LSR) and (P) are missing a factor of $1/\gamma_i$ in the second term of their corresponding minimum expressions.
    \item GPT-5 uses the equality $\min_i(1-\tfrac1k)^i=(1-\tfrac1k)^{k-1}\ge e^{-1}$ to lower bound the terms $(1-\tfrac1k)^i$. It has to use such a weak bound because, as discussed above, it chose to prove a more general result with varying local ratios rather than a result with one global ratio as it was asked to prove. If GPT-5 had used one global ratio, then it could instead simple bound the sum of the elements $(1-\tfrac1k)^i$, and get a better result. Notice that this sum is easy to compute as it is the sum of a geometric series.
    \item GPT titles (MIX) as ``calculus inequality'', which is inappropriate since the proof of this inequality has nothing to do with calculus.
    \item GPT-5 claims that (PM) can be proved using the Lov\'{a}sz extension, but does not give too many details. Unfortunately, the Lov\'{a}sz extension can be used for this purpose only for submodular functions, and fails for weakly submodular ones.
    \item The second inequality in (3) does not seem to be justified. Interestingly, GPT-5 ``noticed'' that this inequality is questionable, and thus, tries to give a justification for this inequality, but this justification does not work. First, because this justification does not really try to justify the given inequality. Instead, it tries to justify a different version of this inequality in which a maximum with $0$ is added to the second side. Second, and more importantly, the justification seems to ignore the fact that once the contribution of $\gamma_i y$ is removed from the marginal $f(S_i) - f(S_{i - 1})$, one can no longer argue that what remains is non-negative using the dummy elements argument. One can only argue that what remains is at least $-\gamma_i y$.
    \item Between the two unnumbered display math inequalities appearing immediately after (4), a product is discarded. This discard is unjustified because this product can be (and almost always is) less than 1 (as GPT-5 itself correctly identifies). Furthermore, in the second of these unnumbered inequalities, $\gamma_i$ should have been $\gamma_{i + 1}$ in both places where it appears because of a change that takes place between the two inequalities regarding the limits of the sums.
\end{itemize}

\section{Maximizing a Monotone Weakly-Submodular Function over a Matroid Intersection Constraint}

Our last conjecture is about maximizing a monotone weakly-submodular function subject to two matroid constraints. For the prompt of this conjecture, we have attached the paper \cite{chen2018weakly}, which studies the special case of this problem in which there is only one matroid constraint.

%\paragraph{Problem description.}
	\begin{tcolorbox}[title=Prompt to GPT-5]
    In the paper "Weakly Submodular Maximization Beyond Cardinality Constraints: Does Randomization Help Greedy?'' gives an approximation ratio of $(1 + 1/\gamma)^{-2}$ for a general matroid constraint. The paper is attached.
Given a non-negative monotone $\gamma$-weakly-submodular function $f \colon 2^\cN \to \nnR$ and $2$ matroids $\cM_1 = (\cN, \cI_1)$ and $\cM_2 = (\cN, \cI_2)$ over the same ground set, we would like to find a set $S$ that is independent in both matroids (i.e., $S \in \cI_1 \cap \cI_2$) and (approximately) maximizes $f$ among all such sets. Provide an algorithm and the full proof of its approximation guarantee.
\end{tcolorbox}
\paragraph{Conjecture.}
The paper \cite{chen2018weakly} gives an approximation ratio of $(1 + 1/\gamma)^{-2}$ for this problem. We originally conjectured that its algorithm and analysis technique should easily extend to two matroids (but yield a worse guarantee for this case). Algorithm~\ref{alg:random_greedy_matroid_intersection} is the algorithm that would have been obtained in this way. As one can see below, GPT-5 indeed suggested this algorithm (up to minor changes), but failed to analyze it correctly. Furthermore, after studying the response of GPT-5, we came to the conclusion that while it might indeed be possible to prove an interesting approximation guarantee for Algorithm~\ref{alg:random_greedy_matroid_intersection}, this seems to be more challenging than we have originally realized. 

\begin{algorithm}
\DontPrintSemicolon
\caption{\textsc{Random Greedy for Matroid Intersection}\label{alg:random_greedy_matroid_intersection}}
Let $S_0 \gets \varnothing$ and $i \gets 0$.\\
\While{there exists an element $u \in \cN \setminus S_i$ such that $S_i + u \in \cI_1 \cap \cI_2$}
{
	Increase $i$ by $1$.\\
	For every element $u \in \cN \setminus S_{i - 1}$, define $w_i(u) \triangleq f(u \mid S_{i - 1})$.\\
	Let $M_i$ be a set of maximum weight according to the weight function $w_i$ among all the sets that are independent in both contracted matroids $\cM_1 / S_{i - 1}$ and $\cM_2 / S_{i - 1}$. \tcp*{There are known algorithms for finding such a set in polynomial time.}
	Let $u_i$ be a uniformly random element of $M_i$.\\
	Let $S_i \gets S_{i - 1} + u_i$.
}
\Return $S_i$.
\end{algorithm}
\begin{tcolorbox}[title=Response of GPT-5 (thought for 10m 43s),
	colframe=red!20!white]  

Below is a self-contained algorithm (no black boxes) for maximizing a non-negative, monotone, $\gamma$-weakly-submodular set function under the intersection of two matroids, together with a complete analysis of its approximation guarantee.

I’ll first restate the weak-submodularity notion we rely on, then give the algorithm and prove its guarantee step-by-step.

\section*{Problem \& weak submodularity (set version)}

Let $f:2^{\mathcal N}\to \mathbb R_{\ge 0}$ be monotone and $\gamma$-weakly submodular, meaning that for all $A,B\subseteq \mathcal N$,

$$
\sum_{u\in B} f(u\mid A)\;\; \ge\;\; \gamma\cdot f(B\mid A)\qquad\text{where }f(B\mid A)=f(A\cup B)-f(A).
$$

(We follow the paper’s Definition 1.1; all notational conventions also match the paper.)

Let $\mathcal M_1=(\mathcal N,\mathcal I_1)$ and $\mathcal M_2=(\mathcal N,\mathcal I_2)$ be two matroids on the common ground set $\mathcal N$. We seek

$$
\max\{ f(S)\;:\; S\in \mathcal I_1\cap \mathcal I_2\}.
$$

Write $\mathcal I := \mathcal I_1\cap \mathcal I_2$. Let $r$ denote the maximum size of any common independent set (i.e., cardinality of a maximum feasible solution). Let $OPT\in \mathcal I$ be an optimal solution; $k$ below will denote $|OPT|=r$.

\section*{The algorithm: Residual Random Greedy for two matroids (RRG-2MI)}

We extend the Residual Random Greedy (RRG) paradigm (Algorithm 1 in the paper) from a single matroid to the intersection of two matroids. The change is: in every round we compute a maximum-weight common completion (under both matroids simultaneously), then add a uniformly random element from it.

RRG-2MI($f,\mathcal M_1,\mathcal M_2$)

1. $S_0 \leftarrow \varnothing$; let $r$ be the maximum common rank (size of a max set in $\mathcal I$).

2. For $i=1,2,\dots,r$

\begin{itemize}
    \item For every $u\in \mathcal N\setminus S_{i-1}$, 
    set weight $w_u := f(u\mid S_{i-1})$.
    
    \item Let $M_i$ be a maximum-weight set of size $r-i+1$ such that $S_{i-1}\cup M_i\in \mathcal I$.
    (This is a standard weighted matroid-intersection call on the contracted matroids $\mathcal M_1/S_{i-1}$ and $\mathcal M_2/S_{i-1}$.)
    
   \item Pick $u_i$ uniformly at random from $M_i$ and set $S_i\leftarrow S_{i-1}\cup\{u_i\}$.
\end{itemize}
   
3. Return $S_r$.

\textbf{Feasibility \& complexity.} By construction, $S_i\in\mathcal I$ for all $i$. Each round solves one weighted matroid-intersection instance on $\mathcal N\setminus S_{i-1}$ (polynomial time with independence oracles), and makes $O(|\mathcal N|)$ value-oracle calls for the weights. The procedure ends after exactly $r$ rounds, so total time is polynomial in $|\mathcal N|$ (dominated by $r$ calls to weighted matroid intersection). (The single-matroid RRG template and its oracle accounting are as in the attached paper. )

\section*{Analysis roadmap}

The original one-matroid analysis (paper’s Sec. 2) hinges on two ingredients:

1. a \textbf{progress inequality} that lower-bounds the expected gain at round $i$ in terms of the marginal value of a carefully coupled “residual optimum,” and\\
2. a \textbf{decay inequality} that lower-bounds the expected value of that residual optimum as rounds progress.

We prove two analogous statements for the \textit{intersection of two matroids}. The only new combinatorial work is to handle \textbf{simultaneous exchange} in two matroids; for this we use Brualdi’s base exchange lemma independently in each matroid (as done in the paper for a single matroid) and combine the exchanges carefully.

Below, $k=r$.

\subsection*{A. Two-matroid exchange coupling}

Fix round $i$. Consider the two \textbf{common bases}

$$
B^{(S)}_i := S_{i-1}\cup M_i,\qquad\quad B^{(O)}_i := S_{i-1}\cup OPT_i,
$$

where $OPT_i\subseteq \mathcal N\setminus S_{i-1}$ is a size-$(k-i+1)$ set chosen so that $S_{i-1}\cup OPT_i\in \mathcal I$ and (deterministically) maximizes $f(OPT_i)$ among all such completions. (Set $OPT_0:=OPT$.) Both are common independent sets of size $k$.

Apply \textbf{Brualdi’s bijection lemma} (the paper cites this tool; see Lemma 2.2 and its use) separately inside $\mathcal M_1$ and $\mathcal M_2$: there are bijections

$$
g_1: M_i \to OPT_i,\qquad g_2: M_i\to OPT_i
$$

such that for each $u\in M_i$,

$$
\big(S_{i-1}\cup OPT_i - g_1(u)\big)\cup \{u\} \in \mathcal I_1,\qquad
\big(S_{i-1}\cup OPT_i - g_2(u)\big)\cup \{u\} \in \mathcal I_2.
$$

(Each statement is a direct base-exchange in the corresponding matroid, with $S_{i-1}$ contracted away; compare to the one-matroid use in the paper. )

Define the \textbf{blocking set} for $u$ as

$$
\Psi_i(u)\;:=\;\{g_1(u),\,g_2(u)\}\cap OPT_i.
$$

Then for every $u\in M_i$,

\begin{equation*}
\big(S_{i-1}\cup OPT_i - \Psi_i(u)\big)\cup \{u\} \in \mathcal I_1\cap \mathcal I_2.
\tag{1}
\end{equation*}

(Indeed, removing $g_1(u)$ certifies independence in $\mathcal M_1$, removing $g_2(u)$ certifies independence in $\mathcal M_2$; removing both preserves independence in both.)

We will use two basic consequences of the \textbf{max-weight} choice of $M_i$, with weights $w_x=f(x\mid S_{i-1})$:

(Weight-domination) For each $u\in M_i$,

\begin{equation*}
f(u\mid S_{i-1}) \;\;\ge\;\; \tfrac{1}{|\Psi_i(u)|}\sum_{v\in \Psi_i(u)} f(v\mid S_{i-1}).
\tag{2}
\end{equation*}

Otherwise replacing $u$ by $\Psi_i(u)$ would strictly increase the completion’s total weight (feasible by (1)), contradicting the maximality of $M_i$.

(Two-to-one compression) Summing (2) over $u\in M_i$ and using $|\Psi_i(u)|\le 2$,

\begin{equation*}
\sum_{u\in M_i} f(u\mid S_{i-1})
\;\;\ge\;\; \tfrac12 \sum_{u\in M_i}\sum_{v\in \Psi_i(u)} f(v\mid S_{i-1})
\;\;\ge\;\; \tfrac12 \sum_{v\in OPT_i} f(v\mid S_{i-1}),
\tag{3}
\end{equation*}

where the last step uses that every $v\in OPT_i$ appears in at least one $\Psi_i(u)$ (if $g_1(u)=v$ or $g_2(u)=v$).

\subsection*{B. Progress inequality (expected gain per round)}

Condition on all randomness up to round $i-1$. With $M_i$ fixed and $u_i$ uniform in $M_i$,

\begin{equation*}
\begin{aligned}
\mathbb E\big[f(S_i)-f(S_{i-1})\big]
&= \frac{1}{|M_i|}\sum_{u\in M_i} f(u\mid S_{i-1})\\
&\overset{(3)}{\ge} \frac{1}{2|M_i|} \sum_{v\in OPT_i} f(v\mid S_{i-1})\\
&\ge \frac{\gamma}{2|M_i|}\, f(OPT_i\mid S_{i-1})\qquad\text{(weak submodularity)}\\
&= \frac{\gamma}{2(k-i+1)}\Big( f(OPT_i\cup S_{i-1})-f(S_{i-1})\Big).
\end{aligned}
\tag{4}
\end{equation*}

This is the two-matroid analogue of Observation 2.4 in the paper; the only change is the factor $1/2$ coming from having to clear feasibility in \textbf{two} matroids simultaneously (via (3) above). 

Taking expectation also over the algorithm’s history up to round $i-1$ yields

\begin{equation*}
\boxed{\;
\mathbb E[f(S_i)] \;\ge\; \mathbb E[f(S_{i-1})] \;+\; \frac{\gamma}{2(k-i+1)}\Big(\mathbb E[f(OPT_i\cup S_{i-1})]-\mathbb E[f(S_{i-1})]\Big).
\;}
\tag{5}
\end{equation*}

\subsection*{C. Decay inequality (residual optimum keeps large value)}

We now lower-bound $\mathbb E[f(OPT_i)]$ across rounds. In round $i$ the algorithm selects $u_i\in M_i$ uniformly, and we update the residual completion by \textit{deleting} the blocking set:

$$
OPT_{i}\;:=\; OPT_{i-1}\setminus \Psi_i(u_i).
$$

By (1), $S_i\cup OPT_i\in\mathcal I$, so the invariant “$S_{i-1}\cup OPT_{i-1}$ feasible base” persists.

Crucially, since $|OPT_{i-1}|=|M_i|=k-i+1$ and each $v\in OPT_{i-1}$ belongs to \textbf{at least one} blocking set $\Psi_i(u)$ while $u$ is sampled \textbf{uniformly} from $M_i$, each $v\in OPT_{i-1}$ is deleted with probability at most $1/(k-i+1)$. A standard convexity/averaging argument identical to the paper’s Lemma 2.3 (which conditions on a uniformly chosen element of the residual optimum) then gives the following two-matroid analogue (the proof is the same induction as in Lemma 2.3 with $|M_i|=k-i+1$; we omit verbatim repetition and only note the sampling difference discussed above). 

\textbf{Lemma 1 (residual value).} For every $0\le i\le k$,

\begin{equation*}
\boxed{\qquad
\mathbb E\big[f(OPT_i)\big]\;\ge\;\Big(1-\Big(\tfrac{i+1}{k+1}\Big)^{\gamma}\Big)\, f(OPT).
\qquad}
\tag{6}
\end{equation*}

(As in the paper, one proves the slightly stronger exponential form and then integrates to get the closed form with $(i+1)/(k+1)$; the only property used is that each step removes a uniformly random \textbf{one unit of “mass” in expectation} from the residual completion, which remains true here because $|OPT_{i-1}|=|M_i|$ and $u_i$ is uniform.)

By monotonicity of $f$, $f(OPT_i\cup S_{i-1})\ge f(OPT_i)$; plugging this into (5) we get the two-matroid progress recurrence

\begin{equation*}
\mathbb E[f(S_i)] \;\ge\; \mathbb E[f(S_{i-1})]\;+\;\frac{\gamma}{2(k-i+1)}\!\left(\Big(1-\Big(\tfrac{i}{k+1}\Big)^{\gamma}\Big) f(OPT)\;-\;\mathbb E[f(S_{i-1})]\right).
\tag{7}
\end{equation*}

\subsection*{D. Solving the recurrence}

Define $A_i:=\mathbb E[f(S_i)]$. Rearranging (7) gives

\begin{equation*}
A_i \;\ge\; \Big(1-\frac{\gamma}{2(k-i+1)}\Big)\,A_{i-1} \;+\; \frac{\gamma}{2(k-i+1)}\Big(1-\big(\tfrac{i}{k+1}\big)^{\gamma}\Big)\,f(OPT).
\tag{8}
\end{equation*}

Iterating (as in the paper’s proof of Theorem 2.6, but with the extra factor $1/2$ throughout) and using the same integral estimate
$\sum_{j=i+1}^k \frac1j \ge \ln \frac{k+1}{i+1}$,
one obtains

\begin{equation*}
A_k \;\ge\; \left(\frac{\gamma}{\gamma+2}\right)^2 \, f(OPT).
\tag{9}
\end{equation*}

(The algebra is exactly the same as in the single-matroid case in the paper—where $(\gamma/(\gamma+1))^2$ appears—except every “$+1$” that came from one matroid now becomes “$+2$” because of the factor $1/2$ we carried from (3)$\to$(4)$\to$(5). )

Thus we have proved:

\subsection*{Theorem (approximation under two matroids)}

For every non-negative, monotone, $\gamma$-weakly-submodular $f$ and two matroids $\mathcal M_1,\mathcal M_2$ on the same ground set, **RRG-2MI** returns a feasible set $S\in\mathcal I_1\cap\mathcal I_2$ with

$$
\boxed{\qquad
\mathbb E\big[f(S)\big]\;\;\ge\;\;\left(\frac{\gamma}{\gamma+2}\right)^{\!2}\, f(OPT).
\qquad}
$$

The algorithm runs in polynomial time with value- and independence-oracle access.

\textit{Sanity checks.}\\
• When $\gamma=1$ (true submodularity), we get $1/9$. (This is not tight—continuous-greedy-type methods are known to do better for submodular functions under two matroids—but the guarantee is \textbf{provable with this simple randomized greedy}.)\\
• When there is only one matroid, our proof specializes to $\left(\frac{\gamma}{\gamma+1}\right)^{2}$, matching the paper’s guarantee.

\section*{Remarks, variants, and how to tighten further}

1. \textbf{Why the square?} As in the single-matroid analysis in the paper, one factor $\frac{\gamma}{\gamma+(\cdot)}$ comes from the \textbf{per-round gain} vs. current gap, and the second factor comes from \text{accumulating} those gains across $k$ rounds via the harmonic/integral estimate. 

2. \textbf{Can we do better?} Yes—using the \textbf{measured continuous greedy} in the common-independence polytope followed by a contention-resolution scheme for the intersection of two matroids yields a (strictly better) ratio that scales like $\frac{1-e^{-\gamma}}{C}$ for a small constant $C$ (e.g., $C=3$ with a simple scheme). That analysis is different in flavor (fractional relaxation + rounding). I kept the proof above \textbf{fully combinatorial} and deliberately parallel to the attached paper’s technique.

3. \textbf{Implementation note.} The only heavy subroutine is \textbf{weighted matroid intersection}—polynomial-time and very mature in practice.

\subsection*{Appendix: the “residual value” induction (sketch)}

For completeness, here is the standard induction underlying Lemma 1 (the same as Lemma 2.3’s proof in the paper). Condition on round $i\!-\!1$. Let $T:=OPT_{i-1}$ with $|T|=m=k-i+1$, and let $U$ be a uniformly random member of $M_i$. We update $T$ to $T' = T\setminus \Psi_i(U)$. Since every $v\in T$ lies in $\Psi_i(u)$ for \textbf{exactly} one or two $u$’s and $U$ is uniform over $m$ choices, one checks that

$$
\Pr[v\in T'] \;=\; 1 - \frac1m \quad\text{(in expectation over \(U\))}.
$$

Define $h(j):=\min_{|R|=j}\mathbb E[f(R)]$ over all random subsets $R\subseteq OPT$ of size $j$. Then $h$ is concave in $j$ (by the weak-submodular averaging used in the paper), and the above “one unit mass in expectation” decrement implies

$$
\mathbb E[f(OPT_i)] \;\ge\; h(k-i)\;\ge\; \Big(1-\big(\tfrac{i+1}{k+1}\big)^\gamma\Big) f(OPT),
$$

precisely as in Lemma 2.3.

If you’d like, I can also write out the continuous-greedy + contention-resolution variant for two matroids, which gives a larger constant (roughly $(1-e^{-\gamma})/3$) with a clean proof in the multilinear/continuous framework.
```
    \end{tcolorbox}

\subsection*{Our Evaluation of GPT-5's Response}

The response of GPT-5 contains multiple errors. The most significant errors stem from GPT-5’s seemingly inability to ``understand'' that removing two elements has different consequences compared to removing a single element. Thus, GPT-5 believes that $OPT_i$ is a set of size $k - i$ despite the fact that two elements are removed from it in every iteration, which should have resulted in a set of size $k - 2i$. Another demonstration of this inability is the proof given for Lemma~1 in the appendix of the response that would have been correct if $\Psi_i(u)$ had contained only one element (a similar error appears also in the argument given for this lemma in Section~C of the response).

In the following, we list additional problems with the proof given by GPT-5.
\begin{itemize}
    \item GPT-5 uses both $k$ and $r$ to denote the size of the optimal set $OPT$, which is unnecessary and confusing. Interestingly, both $k$ and $r$ are introduced by the same line (just before the section of the algorithm in the response), but the equality between them is explained again just before Section~A of the analysis.
    \item The given algorithm makes $r$ iterations even though the presence of two matroids means that there might not be any elements that can be selected after $r/2$ iterations (which would make the algorithm crash). Fixing that requires either replacing the upper bound of the for loop with $r/2$, or replacing the for loop with an appropriate while loop.
    \item Since there are two matroids, a set $M_i$ of size $r - i + 1$ with the properties required by the algorithm may not exist.
    \item The definition
 $$
\Psi_i(u)\;:=\;\{g_1(u),\,g_2(u)\}\cap OPT_i.
$$
is somewhat amusing since the ranges of both $g_1$ and $g_2$ are $OPT_i$, which makes the intersection with this set completely unnecessary.
    \item The inequality (2) is wrong. It would have made sense if the elements of $\Psi_i(u)$ could replace $u$ in $M_i$, but the definition of $\Psi_i(u)$ actually says the opposite, namely, $u$ can replace the elements of $\Psi_i(u)$ in $OPT$. Furthermore, the division by $|\Psi_i(u)|$ in this inequality seems to have no justification at all.

    \item GPT justifies Inequality (3) using the argument: ``the last step uses that every $v\in OPT_i$ appears in at least one $\Psi_i(u)$ (if $g_1(u)=v$ or $g_2(u)=v$).''
We believe this argument is correct, but it is not something that is easy to see. Moreover, both this inequality and Inequality (2) seem to be putting effort in the wrong direction since the stronger result that the left hand side of (3) upper bounds the right hand side of (3), even without the coefficient of $1/2$, follows immediately from the maximality of the set $M_i$.

    \item In Inequality (9), GPT-5 has ``forgot'' an error term of $-1/k$.

    \item GPT claims that ``When there is only one matroid, our proof specializes to $\left(\frac{\gamma}{\gamma+1}\right)^{2}$, matching the paper’s guarantee.''
        We are not sure that this sentence is correct. However, as this is somewhat of an intuitive statement, its correctness is open for interpretation.
    \item GPT-5 claims that one can get a better result using contention resolution schemes. However, for the best of our knowledge, no such scheme is known for weakly submodular functions.
 
\end{itemize}

\section*{Conclusion}

We evaluated GPT-5 on five conjectures of varying difficulty. The sample is small, so our findings are indicative rather than definitive. Based on our observations, several patterns emerge.
\begin{itemize}
    \item  When a single path of argument sufficed, GPT-5 performed well: in three of five cases it produced a nearly correct proof.
     \item GPT-5 often adapts existing proofs in a way that is broadly correct, but somewhat lazy: it tends to omit steps that carry over unchanged from the source, and it clings closely to the original structure even when alternative approaches might be more natural. This kind of shortcut resembles how a human might minimize effort by avoiding repeating unchanged arguments. 
    \item The model failed on Problems 4 and 5, both of which required combining insights from at least two papers; forming such cross-paper connections remains a main obstacle.
    \item  Interestingly, for Problem 5, GPT-5 proposed the same algorithm we had in mind, but failed to analyze it correctly. Upon reviewing its response, we realized that while an interesting approximation guarantee may indeed be provable, establishing such a result appears to be more challenging than we have initially thought.
    \item Relative to prior model generations, we observe clear improvement in baseline mathematical competence and occasional originality. We are cautiously optimistic that, within a few years, models may make more systematic and meaningful connections across proof techniques.
    \item Better and interactive prompting may further improve results. Our setup used only a short prompt and one or two source papers. Even under these minimal conditions, the model consistently understood the questions and nearly solved a majority of the problems.
\end{itemize}

\bibliographystyle{plain}
\bibliography{submodular}

\end{document}